\documentclass{article}
\usepackage{spconf,amsmath,graphicx,hyperref}

\usepackage{booktabs}
\usepackage{multirow}
\usepackage{multicol}
\usepackage{makecell}
\usepackage{amssymb}
\usepackage{listings}
\usepackage{tcolorbox}

\title{When Large Vision-Language Models Meet Person Re-Identification}
\name{Qizao Wang \qquad Bin Li \qquad Xiangyang Xue$^{\dagger}$
\thanks{This work was supported by the National Natural Science Foundation of China under Grant No. 62576110.}
\thanks{$^{\dagger}$Corresponding author.}}
\address{School of Computer Science, Fudan University, Shanghai, China \\
qzwang22@m.fudan.edu.cn, libin@fudan.edu.cn, xyxue@fudan.edu.cn}
\begin{document}
\topmargin=0mm
\ninept
\maketitle
\begin{abstract}
Large Vision-Language Models (LVLMs) that incorporate visual models and large language models have achieved impressive results across cross-modal understanding and reasoning tasks. In recent years, person re-identification (ReID) has also started to explore cross-modal semantics to improve the accuracy of identity recognition. However, effectively utilizing LVLMs for ReID remains an open challenge. While LVLMs operate under a generative paradigm by predicting the next output word, ReID requires the extraction of discriminative identity features to match pedestrians across cameras. In this paper, we propose LVLM-ReID, a novel framework that harnesses the strengths of LVLMs to promote ReID. Specifically, we employ instructions to guide the LVLM in generating one semantic token that encapsulates key appearance semantics from the person image. This token is further refined through our Semantic-Guided Interaction (SGI) module, establishing a reciprocal interaction between the semantic token and visual tokens. Ultimately, the reinforced semantic token serves as the representation of pedestrian identity. Our framework integrates the semantic understanding and generation capabilities of LVLM into end-to-end ReID training, allowing LVLM to capture rich semantic cues during both training and inference. LVLM-ReID achieves competitive results on multiple benchmarks without additional image-text annotations, demonstrating the potential of LVLM-generated semantics to advance person ReID.
\end{abstract}
\section{Introduction}
\label{sec:intro}

\begin{figure}[t!]
    \centering
    \includegraphics[width=0.83\linewidth]{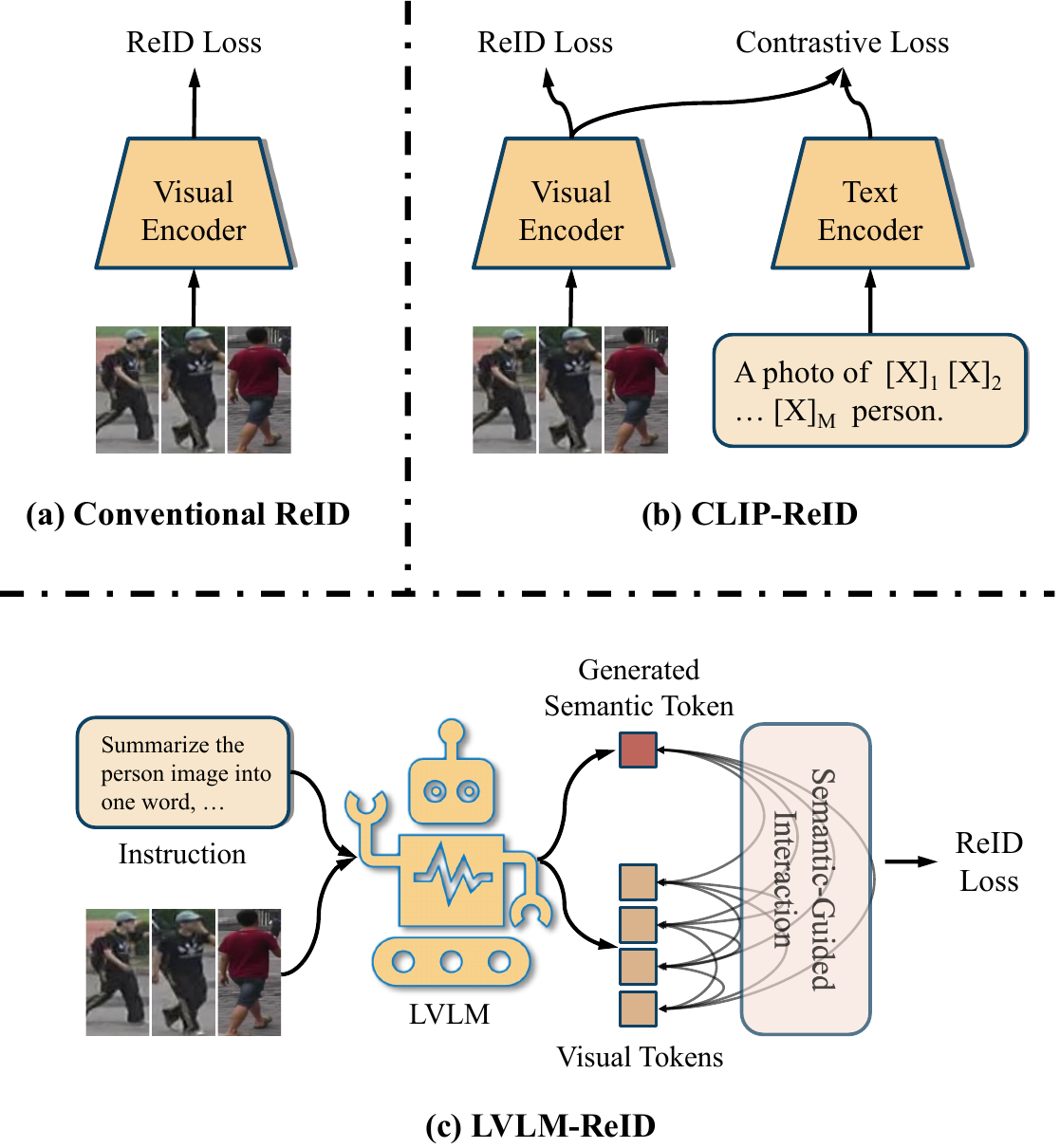}
    \vspace{-0.1in}
    \caption{
    \textbf{Comparison of different person ReID frameworks.}
    }
    \label{fig:intro}
    \vspace{-0.16in}
\end{figure}

\begin{figure*}[!thp]
    \centering
    \includegraphics[width=0.71\linewidth]{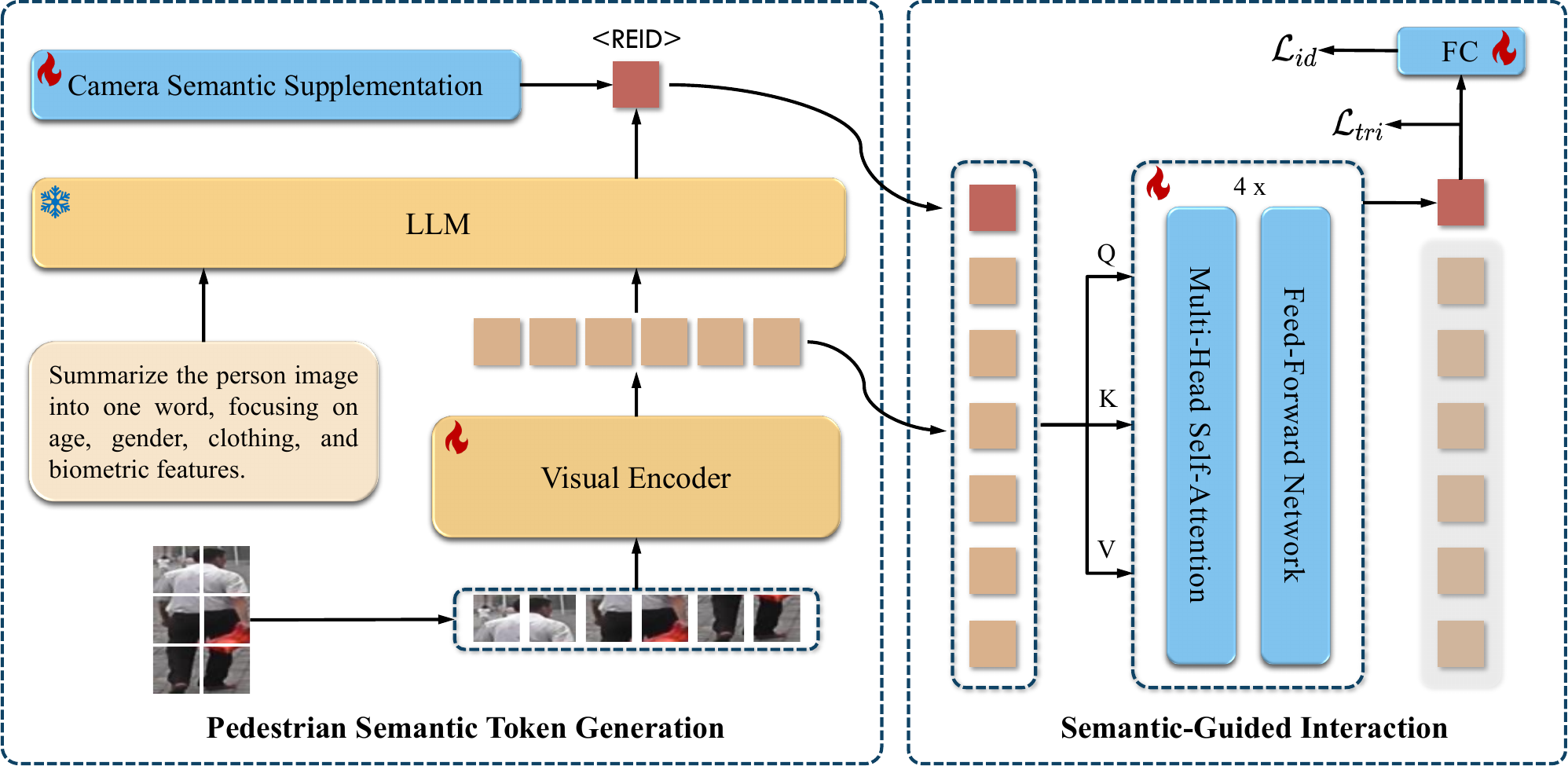}
    \vspace{-0.1in}
    \caption{
    \textbf{Framework of our LVLM-ReID.} 
    It leverages instructions to guide the frozen LLM toward focusing on particular visual semantics within pedestrian images, resulting in the generation of one semantic token that encapsulates the pedestrian's appearance information. Subsequently, an efficient interaction module is designed to facilitate refinement between the generated token and the visual tokens. Finally, the reinforced semantic token is optimized and employed for person retrieval.
    }
    \label{fig:framework}
    \vspace{-0.16in}
\end{figure*}

Person re-identification (ReID) is a crucial task in computer vision, aimed at accurately matching pedestrians across different camera views~\cite{ye2021deep}. With the continuous advancements in deep learning techniques, person ReID methods have evolved significantly~\cite{luo2019strong,wang2023rethinking}. In the past decade, many research has significantly improved ReID accuracy by optimizing the distances between features~\cite{hermans2017defense} and designing refined modules~\cite{jin2020semantics,li2021combined,rao2021counterfactual}, following the paradigm shown in Fig.~\ref{sec:intro}~(a). 
Due to the difficulty of learning rich pedestrian semantics from a single modality, cross-modal learning has received close attention recently. For example, in the context of the development of pre-trained Vision-Language Models (VLMs), CLIP-ReID~\cite{li2023clip} based on the representative VLM model CLIP~\cite{radford2021learning} to leverage text semantics. As shown in Fig.~\ref{sec:intro}~(b), it enhances visual features through cross-modal contrastive learning with image-text pairs. Meanwhile, Large Language Models (LLMs)~\cite{touvron2023llama,yang2024qwen2} have attracted widespread attention due to their powerful capabilities in text generation and comprehension. Large Vision-Language Models (LVLMs)~\cite{achiam2023gpt,liu2024visual,wang2024qwen2} enhance LLMs by incorporating visual perception and understanding, demonstrating considerable potential in multi-modal learning tasks. However, integrating LVLMs with person ReID remains an underexplored challenge.

LVLMs typically operate on a generative paradigm, training and functioning by predicting the next word in a sequence. Thanks to pre-training and instruction tuning, LVLMs can follow instructions and converse with humans. As a result, a direct approach might be to have the model to identify the input person images. However, ReID gallery databases are usually very large (comprising tens of thousands of pedestrian images)~\cite{market1501,ristani2016MTMC}. The time and cost of comparing identities one by one with LVLMs are substantial. Processing multiple images simultaneously would also lead to an unacceptable increase in visual tokens.
Therefore, we are motivated to leverage the reasoning and understanding capabilities of LVLMs, while adhering to the mainstream ReID paradigm of feature extraction combined with feature similarity-based retrieval~\cite{ye2021deep}. 
A potential solution involves using LVLMs to describe pedestrian images and fine-tuning the visual encoder via tasks such as image-text matching or image caption prediction.
However, it presents several limitations:
(1) High-quality and diverse text annotations are expensive to obtain.
(2) The goals of image-text matching or image caption prediction tasks may not align well with those of image-based ReID.
(3) During the inference phase, the potential of LVLMs is often underutilized, as they are not effectively integrated with the visual features.

To address these issues, we propose a new ReID framework called LVLM-ReID to leverage the superior semantic understanding and generation ability of LVLMs. Specifically, as shown in Fig.~\ref{fig:intro}~(c), we use instruction to guide the LVLM to focus on specific visual semantics in pedestrian images, generating a semantic token representing the pedestrian’s appearance information. We then design an effective interaction module between the generated token and visual tokens, refining the visual representations of pedestrians while reinforcing the semantic token as a discriminative identity representation. Ultimately, the reinforced semantic token is optimized and used during inference to achieve person retrieval. Our framework integrates the generative process of LVLMs into the ReID model, eliminating the need for additional image caption annotations and enabling end-to-end effective learning. More importantly, during the inference phase, we continue to leverage the generative power of LVLMs to enhance visual features adaptively. Our experiments show that one LVLM-generated semantic token can effectively facilitate the learning of pedestrian representations.
\textbf{Our contributions are summarized as follows:}
\textbf{(1)} We propose a novel framework that incorporates LVLMs into the person ReID task, offering a new perspective on using generative language models to assist discriminative visual models.
\textbf{(2)} We propose to utilize the generative capability of LVLMs to produce a semantic token for pedestrians and design a semantic-guided interaction module leveraging the generated semantic token to enhance identity representations.
\textbf{(3)} Experimental results show that, without requiring additional annotations, our method effectively improves the discriminability of identity features and achieves competitive results across multiple datasets.

\section{Methodology}
\label{sec:method}

\subsection{Overview of LVLM}
\label{subsec:overview}
\noindent \textbf{Overall framework.}
A typical LVLM consists of three key components: a visual encoder, a vision-language connector, and an LLM. The visual encoder extracts rich visual representations from images, which are then processed by the vision-language connector that converts visual features into the word embedding space. The LLM, trained for next-word prediction, generates text based on the encoded visual content. 
In this work, we leverage Qwen2-VL~\cite{wang2024qwen2}, one of the most advanced LVLMs, known for its superior capabilities in instruction-following, semantic understanding, and text generation across diverse tasks. Qwen2-VL combines a Vision Transformer (ViT)~\cite{dosovitskiy2020image} as the visual encoder and the Qwen2~\cite{yang2024qwen2} as the LLM. The vision-language connector between the two components is one MLP layer that also compresses the extracted visual tokens.

\noindent \textbf{Visual token extraction.}
Before inputting a pedestrian image into the LLM, the image is first encoded and compressed by the visual encoder. Specifically, each input RGB image $x \in \mathbb{R}^{H \times W \times 3}$, where $H$ and $W$ are its height and width, is first divided into patches of size $P \times P$. These patches are then embedded and flattened into a feature vector $x^{p} \in \mathbb{R}^{N \times d}$, where $N=H \times W / P^2$ represents the number of patches, and $d$ is the embedding dimension. The resulting patch embeddings are processed through multiple layers of Transformer self-attention blocks~\cite{vaswani2017attention}, producing visual representations $f \in \mathbb{R}^{N \times d}$. To enhance the model’s ability to capture spatial dependencies, Multimodal Rotary Position Embedding (M-RoPE)~\cite{wang2024qwen2} is used in the process. Afterward, a simple MLP layer compresses adjacent $2 \times 2$ tokens into a single token, producing the final visual tokens $V$, which is formulated as:
\begin{equation}
    V = {\rm MLP}(f) = \left[v_{1}; \ v_{2}; \ \cdots; \ v_{n} \right] \in \mathbb{R}^{n \times d},
\end{equation}
where $n = \frac{N}{4}$. Notably, instead of using the traditional \texttt{[class]} token~\cite{dosovitskiy2020image}, the image is transformed into a set of visual tokens. These visual tokens will then be passed to the LLM for further processing and interaction.

\subsection{Pedestrian Semantic Token Generation}
\label{subsec:PSTG}
We aim to integrate the advanced visual semantic understanding and generation capabilities of LVLM into the feature extraction pipeline, by guiding the ReID model to generate one semantic token that encapsulates instructive information of the pedestrian. 
To achieve this, we use instructions to direct the LVLM to summarize the pedestrian’s visual appearance:
\begin{tcolorbox}[colback=black!5!white,colframe=black!75!black] \texttt{\small <|vision\_start|>} $V$ \texttt{\small <|vision\_end|>} Summarize the person image into one word, focusing on age, gender, clothing, and biometric features. 
\end{tcolorbox}
\noindent where $V$ represents the extracted visual tokens, while the special tokens \texttt{\small <|vision\_start|>} and \texttt{\small <|vision\_end|>} are used to mark the beginning and end of the visual token sequence. With this instruction, the LVLM is guided to focus on the appearance-related semantics in the image, and then generate a semantic token that summarizes the relevant identity features. We denote this generated token as \texttt{<REID>}, which serves as a compact representation of the pedestrian's visual appearance.

\noindent \textbf{Camera semantic supplementation.} 
The semantic token generation process overlooks the influence of camera variations. To improve pedestrian semantic consistency across cameras, we explicitly model and account for these camera-induced feature variations. 
Specifically, we assign a unique learnable embedding vector to each camera, which allows the model to learn the inherent feature shifts caused by cameras. 
We denote the set of learnable camera embeddings as $V_{cam} = \{ v^{i}_{cam}|i = 1, 2, \cdots , N^{c} \}$, where $N^{c}$ is the total number of cameras. One direct implementation is to supplement the generated pedestrian semantic token with the camera semantics as follows:
\begin{equation}
    \bar{v}_{reid} = v_{reid} + v^{y^{c}}_{cam},
\end{equation}
where $v_{reid}$ is the encoding of the \texttt{<REID>} token, $y^{c}$ is the camera ID corresponding to the image $x$. However, this late supplementation strategy may affect the visual model weakly.
We thus try to transfer the usage of camera embeddings to the input of the visual model, where the camera embeddings are added to the patch embeddings $x^p$. We evaluate and discuss the two variants in Sec.~\ref{subsec:ablation}.

\subsection{Semantic-Guided Interaction}
\label{subsec:SGI}
We design the Semantic-Guided Interaction (SGI) module to facilitate bidirectional interaction between the generated semantic token and the visual tokens.
Specifically, the generated semantic token is first concatenated with the visual tokens.
Formally, 
\begin{equation}
    z = \left[{v}_{reid}; \ v_{1}; \ v_{2}; \ \cdots; \ v_{n} \right] \in \mathbb{R}^{(n+1) \times d}.
\end{equation}

This concatenated token sequence is then passed through 4 layers of Transformer blocks, each consisting of a multi-head self-attention layer and a feed-forward network. 
The module refines the visual features to capture identity-relevant information under the guidance of the semantic token. 
Meanwhile, the semantic token, serving as the pivot for information aggregation, distills more discriminative features from the visual representations, enhancing the overall understanding of the pedestrian’s identity. 
Through the semantic-guided interaction module, the model produces the reinforced representation as:
\begin{equation}
    \hat{z} = \left[\hat{v}_{reid}; \ \hat{v}_{1}; \ \hat{v}_{2}; \ \cdots; \ \hat{v}_{n} \right] = {\rm SGI}(z).
\end{equation}

Then, the reinforced semantic token representation $\hat{v}_{reid}$ is used to compute the Re-ID losses, \textit{i.e.}, identity classification loss~\cite{luo2019strong} and triplet loss~\cite{hermans2017defense}.

\subsection{Optimization and Inference}
\label{subsec:optim-infer}
During training, we optimize the parameters of both the visual model and the SGI module. The parameters of LLM are frozen, but gradients backpropagate through it to update other learnable components.
By leveraging the generated \texttt{<REID>} token in conjunction with the SGI module, we achieve a joint end-to-end training that harnesses the strengths of LVLM in instruction-following and visual semantic understanding. This process allows for the integration of rich semantic cues into the visual representations, improving the accuracy of pedestrian identity recognition.
The overall training loss is a weighted combination of the identity classification loss $\mathcal{L}_{id}$~\cite{luo2019strong} and the triplet loss $\mathcal{L}_{tri}$~\cite{hermans2017defense}, which is expressed as follows:
\begin{equation}
    \mathcal{L} = \alpha_1 \mathcal{L}_{id} + \alpha_2 \mathcal{L}_{tri},
\label{eq:overall_loss}
\end{equation}
where $\alpha_1$ and $\alpha_2$ are balancing factors that control the contribution of each loss term. 

During inference, the LVLM is also used to generate the \texttt{<REID>} token for each input image. Then, the reinforced semantic token representation, $\hat{v}_{reid}$, is used to compute the cosine similarity between different person images. These similarity scores are employed for identity matching, allowing the model to identify pedestrians. Note that the identity representations of persons in the large gallery databases need be extracted only once in applications.

\begin{table}[t]
  \centering
  \caption{\textbf{Comparison with the state-of-the-art methods.} The results of our method and the best results of comparison methods are shown in bold.}
  \label{tab:compare_SOTA}
  \resizebox{1\linewidth}{!}{
  \begin{tabular}{llcccccccc}
    \toprule
    \multirow{2}[0]{*}{\textbf{Backbone}} & \multirow{2}[0]{*}{\textbf{Methods}} & \multicolumn{2}{c}{\textbf{DukeMTMC-reID}} & \multicolumn{2}{c}{\textbf{Market-1501}} & \multicolumn{2}{c}{\textbf{CUHK03}} & \multicolumn{2}{c}{\textbf{Occluded-Duke}} \\
    \cmidrule(r){3-4} \cmidrule(r){5-6} \cmidrule(r){7-8} \cmidrule(r){9-10}
    & & mAP & Rank-1 & mAP & Rank-1 & mAP & Rank-1 & mAP & Rank-1 \\
    \midrule
    \multirow{8}{*}{CNN} & MGN~\cite{wang2018learning} & 78.4 & 88.7 & 86.9 & 95.7 & 67.4 & 68.0 & - & - \\
    & DG-Net~\cite{zheng2019joint} & 74.8 & 86.6 & 86.0 & 94.8 & - & - & - & - \\
    & SAN~\cite{jin2020semantics} & 75.5 & 87.9 & 88.0 & \bf 96.1 & 76.4 & 80.1 & - & - \\
    & Pyramid~\cite{zheng2019pyramidal} & 79.0 & 89.0 & 88.2 & 95.7 & 76.9 & 78.9 & - & - \\
    & Relation-Net~\cite{park2020relation} & 78.6 & 89.7 & 88.9 & 95.2 & 75.6 & 77.9 & - & - \\
    & RGA-SC~\cite{zhang2020relation} & - & - & 88.4 & \bf 96.1 & 77.4 & 81.1 & - & - \\
    & CDNet~\cite{li2021combined} & 76.8 & 88.6 & 86.0 & 95.1 & - & - & - & - \\ 
    & CAL~\cite{rao2021counterfactual} & 76.4 & 87.2 & 87.0 & 94.5 & - & - & - & - \\
    \midrule
    
    \multirow{5}{*}{ViT} & TransReID~\cite{he2021transreid} & 80.6 & 89.6 & 88.2 & 95.0 & - & - & 55.7 & 64.2 \\
    & PAT~\cite{li2021diverse} & 78.2 & 88.8 & 88.0 & 95.4 & - & - & 53.6 & 64.5 \\
    & DCAL~\cite{zhu2022dual} & 80.1 & 89.0 & 87.5 & 94.7 & - & - & - & - \\
    & AAformer~\cite{zhu2023aaformer} & 80.0 & \bf 90.1 & 88.0 & 95.4 & 79.0 & 80.3 & 58.2 & \bf 67.1 \\
    & CLIP-ReID~\cite{li2023clip} & \bf 82.5 & 90.0 & \bf 89.6 & 95.5 & \bf 80.3 & \bf 81.6 & \bf 59.5 & \bf 67.1 \\
    \midrule
    
    \multicolumn{1}{l}{LVLM} & LVLM-ReID (Ours) & \textbf{82.8} & \textbf{92.2} & \textbf{89.2} & \textbf{95.6} & \textbf{82.3} & \textbf{84.6} & \textbf{59.8} & \textbf{68.1} \\
    \bottomrule
  \end{tabular}}
  \vspace{-0.1in}
\end{table}

\section{Experiments}

\subsection{Experimental Settings}
\label{subsec:exp_settings}
\noindent \textbf{Datasets.}
We evaluate our methods on four person ReID datasets: DukeMTMC-reID~\cite{ristani2016MTMC}, Market-1501~\cite{market1501}, CUHK03~\cite{li2014deepreid}, and Occluded-Duke~\cite{miao2019pose}.

\noindent \textbf{Evaluation metrics.}
We follow the common practices to adopt Cumulative Matching Characteristics (CMC) at Rank-1 and mean Average Precision (mAP) for performance evaluation.

\noindent \textbf{Implementation details.}
We employ Qwen2-VL-2B~\cite{wang2024qwen2} considering its efficiency with limited resources, while larger model sizes such as 7B and 72B have better LLM capabilities. The model adopts BFloat16 mixed precision. $H$, $W$, and $P$ are set to 280, 140, and 14, respectively, resulting in $n=50$. In other words, 50 visual tokens are included in the input of LLM and our SGI module. Following~\cite{luo2019strong}, random horizontal flipping, padding, random cropping, and random erasing~\cite{zhong2020random} are used for data augmentation. 16 identities and 4 images per person are randomly sampled to constitute a training batch. Adam optimizer with weight decay of $3 \times 10^{-4}$ is adopted, with the warmup strategy that linearly increases the learning rate from $3 \times 10^{-5}$ to $3 \times 10^{-4}$ in the first 10 epochs. We train the model for 60 epochs, with a learning rate decay factor of 0.1 at the 30th epoch. $\alpha_1$ and $\alpha_2$ are set to 0.25 and 1 following~\cite{li2023clip}. The margin $m$ of triplet loss is set to 0.3. Our method is implemented using PyTorch and on one NVIDIA A800 GPU.

\subsection{Comparison with State-of-the-Art Methods}
As shown in Tab.~\ref{tab:compare_SOTA}, methods based on CNNs achieve solid performance by designing elaborate modules for person ReID, while TransReID~\cite{he2021transreid} explores the potential of Transformer~\cite{vaswani2017attention,dosovitskiy2020image} in ReID, establishing itself as a strong baseline with superior capability. 
Rather than designing elaborate modules for interactions between image pairs~\cite{zhu2022dual}, or leveraging part-level features~\cite{zhu2023aaformer,li2021diverse} based on ViT, we introduce LVLM's advanced understanding and generative processes into the ReID framework. 
On the DukeMTMC-reID dataset, which is known for variations in appearance, LVLM-ReID achieves an mAP of 82.8\% and a Rank-1 accuracy of 92.2\%, surpassing previous advanced methods.
On the CUHK03 dataset, LVLM-ReID also significantly outperforms the advanced method CLIP-ReID~\cite{li2023clip}. 
LVLM-ReID achieves competitive results on the Market-1501 dataset and also performs well on the challenging Occluded-Duke dataset, showing its robustness and generalization ability in occlusion scenarios.
Note that CLIP-ReID~\cite{li2023clip} leverages a VLM pre-trained on large-scale image-text pairs, and it discards the text encoder during inference.
Differently, LVLM-ReID integrates LVLM into ReID training and inference stages in a novel paradigm. The strong performance of LVLM-ReID across datasets demonstrates its capability as a powerful LVLM-based baseline.

\subsection{Ablation Studies}
\label{subsec:ablation}

\begin{table}[t]
\begin{minipage}[t]{0.45\linewidth}
\centering
\caption{\label{tab:ablation_module}\textbf{Ablation studies of our key two components on DukeMTMC-reID and Market-1501.} }
\vspace{10pt}
\resizebox{1\linewidth}{!}{
\begin{tabular}{lcccc}
\toprule
\multirow{2}{*}{\bf Methods} & \multicolumn{2}{c}{\bf DukeMTMC-reID} & \multicolumn{2}{c}{\bf Market-1501} \\ 
    \cmidrule(r){2-3} \cmidrule(r){4-5}
    & mAP & Rank-1 & mAP & Rank-1 \\ 
    \midrule
    
    Baseline & 79.0 & 90.2 & 87.3 & 94.7 \\
    Ours \textit{w/o} PSTG & 80.9 & 91.0 & 88.3 & 95.0 \\
    Ours \textit{w/o} SGI & 79.0 & 90.0 & 87.3 & 94.5 \\
    \midrule
    Ours & \bf 82.8 & \bf 92.2 & \bf 89.2 & \bf 95.6 \\
    \bottomrule
\end{tabular}}
\end{minipage} \hfill
\begin{minipage}[t]{0.49\linewidth}
\centering
\caption{\label{tab:ablation_css}\textbf{Ablation of the camera semantic supplementation (CSS) strategy.} CSS-$v_{reid}$ and CSS-$x^p$ denote adding the camera embedding to $v_{reid}$ and $x^p$, respectively. }
\resizebox{1\linewidth}{!}{
\begin{tabular}{lcccc}
\toprule
\multirow{2}{*}{\bf Methods} & \multicolumn{2}{c}{\bf DukeMTMC-reID} & \multicolumn{2}{c}{\bf Market-1501} \\ 
    \cmidrule(r){2-3} \cmidrule(r){4-5}
    & mAP & Rank-1 & mAP & Rank-1 \\ 
    \midrule

    \textit{w/o} CSS & 81.6 & 91.4 & 89.1 & 95.2 \\
    CSS-$v_{reid}$ & 82.3 & 92.1 & 88.4 & 95.3 \\
    CSS-$x^p$ & \bf 82.8 & \bf 92.2 & \bf 89.2 & \bf 95.6 \\
    \bottomrule
\end{tabular}}
\end{minipage}
\vspace{-0.1in}
\end{table}

\noindent \textbf{Effectiveness of the generated pedestrian semantic token.}
(1) Our baseline only uses the visual model of the LVLM, and the visual tokens are averaged to compute loss and feature similarity during training and inference. 
The baseline only uses the visual model, overlooking the role of LVLM in visual semantic understanding and achieving inferior performance. 
(2) In the variant ``Ours w/o PSTG'', we replace the LVLM-generated semantic token with a learnable token, similar to the design of the \texttt{[class]} token~\cite{dosovitskiy2020image}, to integrate visual information. As shown in Tab.~\ref{tab:ablation_module}, this substitution leads to a substantial performance drop since the randomly initialized learnable token lacks rich semantic cues. This result underscores the importance of our PSTG mechanism, which contributes to a more comprehensive understanding of pedestrian images.

\noindent \textbf{Effectiveness of the SGI module.}
In the ``Ours \textit{w/o} SGI'' variant, we remove the SGI module and rely solely on the LVLM-generated semantic token for ReID. As shown in Tab.~\ref{tab:ablation_module}, this configuration achieves reasonably good performance, suggesting that our PSTG effectively captures essential pedestrian semantic information. However, the variant struggles to outperform the baseline, emphasizing the importance of the SGI module in leveraging the generated semantic token.
It represents a novel paradigm of enhancing identity representations with the LVLM-generated semantic token.

\noindent \textbf{Ablation of the camera semantic supplementation strategy.}
In Tab.~\ref{tab:ablation_css}, the result of CSS-$v_{reid}$ shows that camera semantics can improve the representation ability of the generated token. 
When transferring the usage of camera embeddings to the input of the visual model (denoted by CSS-$x^p$), we observe a better performance. 
This design helps to improve the robustness of visual features and the LVLM-generated semantic token, further improving the model's ability to match pedestrians across cameras. 

\begin{figure}[t]
    \centering
    \includegraphics[width=0.85\linewidth]{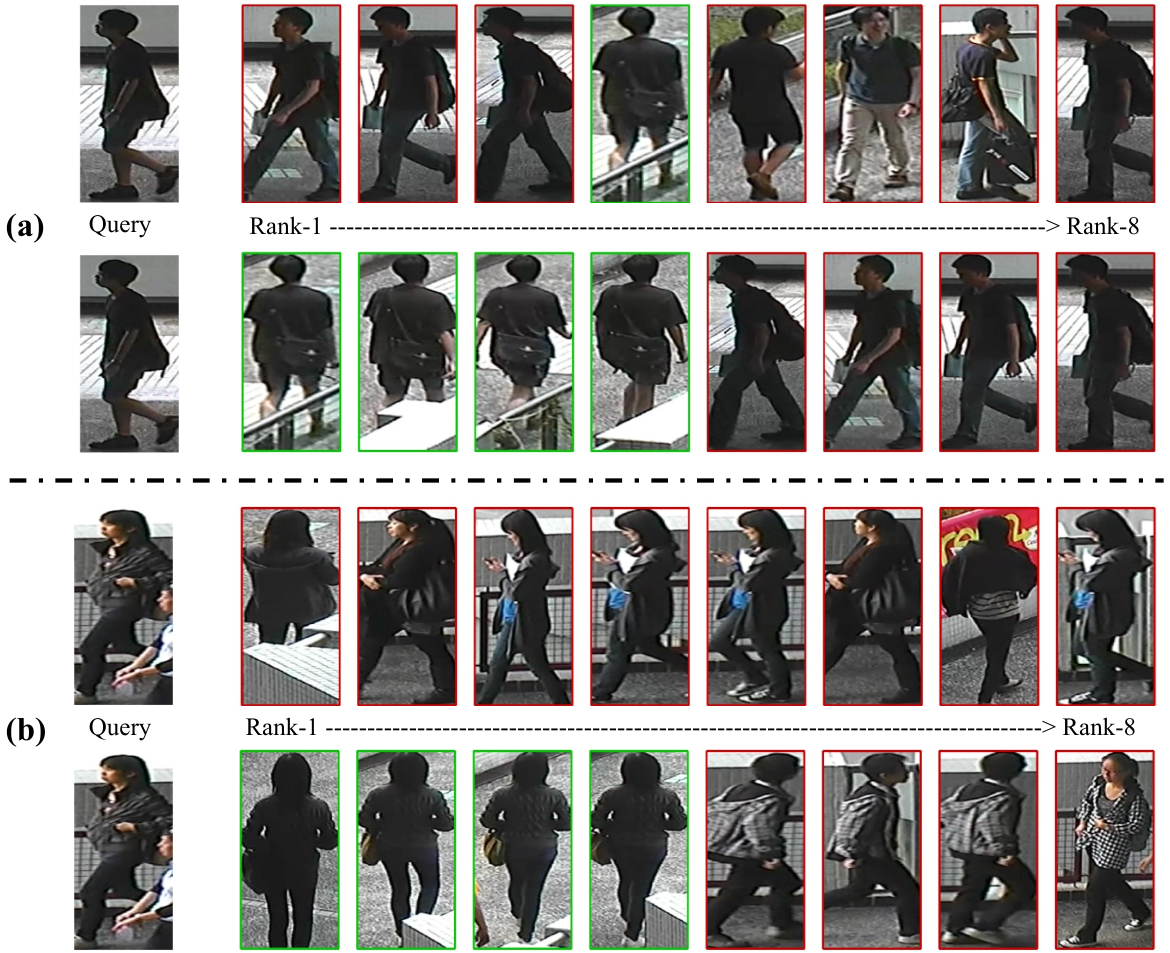}
    \vspace{-0.1in}
    \caption{
    \textbf{Visualization of retrieval results.} For each query, the first and the second rows show the top-8 retrieval results of the baseline and our method on CUHK03, respectively. Retrieved images with green and red boxes are correct and incorrect results, respectively. For each query in the figure, there are four matched person images in the gallery. Best viewed in color and zoomed in.
    }
    \label{fig:cuhk_retrieval}
    \vspace{-0.1in}
\end{figure}

\noindent \textbf{Inference time.}
Our method requires generating only a single semantic token using LVLM, which can be accomplished in a single forward pass during inference, eliminating the need for iterative autoregressive generation while ensuring practicality.
The inference time for processing 64 images in parallel is 784 ms for the visual-only baseline and 906 ms for our model on a single GPU with 12GB of memory. FLOPs are 128G and 257G, respectively. The marginal increase in inference time is acceptable and can be further optimized through techniques such as flash attention~\cite{dao2023flashattention}.

\noindent \textbf{Visualization of retrieval results.}
As shown in Fig.~\ref{fig:cuhk_retrieval}, the baseline model often relies on coarse-grained and vulnerable features, such as general color, similar clothing, and similar pose, leading to false positives. 
In contrast, guided by the semantic token, our method effectively captures nuanced identity-specific features, such as distinctive patterns and accessories, enabling superior differentiation between individuals.
Our method also showcases robustness in scenarios involving occlusions, whether from objects (Fig.~\ref{fig:cuhk_retrieval}~(a)) or other pedestrians (Fig.~\ref{fig:cuhk_retrieval}~(b)).

\section{Conclusion}
\label{sec:conclusion}
In this paper, we introduce LVLM-ReID, a novel framework that leverages the semantic understanding and generation capabilities of LVLMs to promote person ReID. We certify that LVLM can be integrated into the ReID process by generating one pedestrian semantic token, which is then used for efficient interaction with visual tokens. 
In our framework, LVLM effectively helps capture and utilize the rich semantics of pedestrians. 
Experimental results show the significance of LVLM-based semantic guidance in enhancing visual representations, paving the way for future research in this direction.

\noindent \textbf{Future work.} 
Misleading the LVLM to attend to background reduces Rank-1 by 0.6\% on CUHK03, and LVLM-ReID underperfoms under heavy occlusion or low resolution. The adverse effects of inaccurate semantics and LVLM hallucination~\cite{liu2024survey} as well as degradation in challenging scenarios point to future directions such as better instructions, multiple semantic tokens, or larger LVLM variants.

{
\fontsize{8.82pt}{5pt}\selectfont
\bibliographystyle{IEEEbib}
\bibliography{refs}

\begin{thebibliography}{10}

\bibitem{ye2021deep}
Mang Ye, Jianbing Shen, Gaojie Lin, Tao Xiang, Ling Shao, and Steven~CH Hoi,
\newblock ``Deep learning for person re-identification: A survey and outlook,''
\newblock {\em IEEE Transactions on Pattern Analysis and Machine Intelligence}, vol. 44, no. 6, pp. 2872--2893, 2021.

\bibitem{luo2019strong}
Hao Luo, Wei Jiang, Youzhi Gu, Fuxu Liu, Xingyu Liao, Shenqi Lai, and Jianyang Gu,
\newblock ``A strong baseline and batch normalization neck for deep person re-identification,''
\newblock {\em IEEE Transactions on Multimedia}, vol. 22, no. 10, pp. 2597--2609, 2019.

\bibitem{wang2023rethinking}
Qizao Wang, Xuelin Qian, Bin Li, Yanwei Fu, and Xiangyang Xue,
\newblock ``Rethinking person re-identification from a projection-on-prototypes perspective,''
\newblock {\em arXiv preprint arXiv:2308.10717}, 2023.

\bibitem{hermans2017defense}
Alexander Hermans, Lucas Beyer, and Bastian Leibe,
\newblock ``In defense of the triplet loss for person re-identification,''
\newblock {\em arXiv preprint arXiv:1703.07737}, 2017.

\bibitem{jin2020semantics}
Xin Jin, Cuiling Lan, Wenjun Zeng, Guoqiang Wei, and Zhibo Chen,
\newblock ``Semantics-aligned representation learning for person re-identification,''
\newblock in {\em Proceedings of the AAAI Conference on Artificial Intelligence}, 2020, vol.~34, pp. 11173--11180.

\bibitem{li2021combined}
Hanjun Li, Gaojie Wu, and Wei-Shi Zheng,
\newblock ``Combined depth space based architecture search for person re-identification,''
\newblock in {\em Proceedings of the IEEE/CVF Conference on Computer Vision and Pattern Recognition}, 2021, pp. 6729--6738.

\bibitem{rao2021counterfactual}
Yongming Rao, Guangyi Chen, Jiwen Lu, and Jie Zhou,
\newblock ``Counterfactual attention learning for fine-grained visual categorization and re-identification,''
\newblock in {\em Proceedings of the IEEE/CVF International Conference on Computer Vision}, 2021, pp. 1025--1034.

\bibitem{li2023clip}
Siyuan Li, Li~Sun, and Qingli Li,
\newblock ``Clip-reid: exploiting vision-language model for image re-identification without concrete text labels,''
\newblock in {\em Proceedings of the AAAI Conference on Artificial Intelligence}, 2023, vol.~37, pp. 1405--1413.

\bibitem{radford2021learning}
Alec Radford, Jong~Wook Kim, Chris Hallacy, Aditya Ramesh, Gabriel Goh, Sandhini Agarwal, Girish Sastry, Amanda Askell, et~al.,
\newblock ``Learning transferable visual models from natural language supervision,''
\newblock in {\em International Conference on Machine Learning}, 2021, pp. 8748--8763.

\bibitem{touvron2023llama}
Hugo Touvron, Thibaut Lavril, Gautier Izacard, Xavier Martinet, Marie-Anne Lachaux, Timoth{\'e}e Lacroix, Baptiste Rozi{\`e}re, Naman Goyal, et~al.,
\newblock ``Llama: Open and efficient foundation language models,''
\newblock {\em arXiv preprint arXiv:2302.13971}, 2023.

\bibitem{yang2024qwen2}
An~Yang, Baosong Yang, Binyuan Hui, Bo~Zheng, Bowen Yu, Chang Zhou, Chengpeng Li, Chengyuan Li, et~al.,
\newblock ``Qwen2 technical report,''
\newblock {\em arXiv preprint arXiv:2407.10671}, 2024.

\bibitem{achiam2023gpt}
Josh Achiam, Steven Adler, Sandhini Agarwal, Lama Ahmad, Ilge Akkaya, Florencia~Leoni Aleman, Diogo Almeida, Janko Altenschmidt, Sam Altman, Shyamal Anadkat, et~al.,
\newblock ``Gpt-4 technical report,''
\newblock {\em arXiv preprint arXiv:2303.08774}, 2023.

\bibitem{liu2024visual}
Haotian Liu, Chunyuan Li, Qingyang Wu, and Yong~Jae Lee,
\newblock ``Visual instruction tuning,''
\newblock {\em Advances in Neural Information Processing Systems}, vol. 36, 2024.

\bibitem{wang2024qwen2}
Peng Wang, Shuai Bai, Sinan Tan, Shijie Wang, Zhihao Fan, Jinze Bai, Keqin Chen, Xuejing Liu, et~al.,
\newblock ``Qwen2-vl: Enhancing vision-language model's perception of the world at any resolution,''
\newblock {\em arXiv preprint arXiv:2409.12191}, 2024.

\bibitem{market1501}
Liang Zheng, Liyue Shen, Lu~Tian, Shengjin Wang, Jingdong Wang, and Qi~Tian,
\newblock ``Scalable person re-identification: A benchmark,''
\newblock in {\em Proceedings of the IEEE/CVF International Conference on Computer Vision}, 2015, pp. 1116--1124.

\bibitem{ristani2016MTMC}
Ergys Ristani, Francesco Solera, Roger Zou, Rita Cucchiara, and Carlo Tomasi,
\newblock ``Performance measures and a data set for multi-target, multi-camera tracking,''
\newblock in {\em Proceedings of the European Conference on Computer Vision Workshops}, 2016, pp. 17--35.

\bibitem{dosovitskiy2020image}
Alexey Dosovitskiy, Lucas Beyer, Alexander Kolesnikov, Dirk Weissenborn, Xiaohua Zhai, Thomas Unterthiner, Mostafa Dehghani, Matthias Minderer, Georg Heigold, Sylvain Gelly, et~al.,
\newblock ``An image is worth 16x16 words: Transformers for image recognition at scale,''
\newblock in {\em Proceedings of the International Conference on Learning Representations}, 2021.

\bibitem{vaswani2017attention}
Ashish Vaswani, Noam Shazeer, Niki Parmar, Jakob Uszkoreit, Llion Jones, Aidan~N Gomez, {\L}ukasz Kaiser, and Illia Polosukhin,
\newblock ``Attention is all you need,''
\newblock in {\em Advances in Neural Information Processing Systems}, 2017, pp. 5998--6008.

\bibitem{wang2018learning}
Guanshuo Wang, Yufeng Yuan, Xiong Chen, Jiwei Li, and Xi~Zhou,
\newblock ``Learning discriminative features with multiple granularities for person re-identification,''
\newblock in {\em Proceedings of the 26th ACM International Conference on Multimedia}, 2018, pp. 274--282.

\bibitem{zheng2019joint}
Zhedong Zheng, Xiaodong Yang, Zhiding Yu, Liang Zheng, Yi~Yang, and Jan Kautz,
\newblock ``Joint discriminative and generative learning for person re-identification,''
\newblock in {\em Proceedings of the IEEE/CVF Conference on Computer Vision and Pattern Recognition}, 2019, pp. 2138--2147.

\bibitem{zheng2019pyramidal}
Feng Zheng, Cheng Deng, Xing Sun, Xinyang Jiang, Xiaowei Guo, Zongqiao Yu, Feiyue Huang, and Rongrong Ji,
\newblock ``Pyramidal person re-identification via multi-loss dynamic training,''
\newblock in {\em Proceedings of the IEEE/CVF Conference on Computer Vision and Pattern Recognition}, 2019, pp. 8514--8522.

\bibitem{park2020relation}
Hyunjong Park and Bumsub Ham,
\newblock ``Relation network for person re-identification,''
\newblock in {\em Proceedings of the AAAI Conference on Artificial Intelligence}, 2020, vol.~34, pp. 11839--11847.

\bibitem{zhang2020relation}
Zhizheng Zhang, Cuiling Lan, Wenjun Zeng, Xin Jin, and Zhibo Chen,
\newblock ``Relation-aware global attention for person re-identification,''
\newblock in {\em Proceedings of the IEEE/CVF Conference on Computer Vision and Pattern Recognition}, 2020, pp. 3186--3195.

\bibitem{he2021transreid}
Shuting He, Hao Luo, Pichao Wang, Fan Wang, Hao Li, and Wei Jiang,
\newblock ``Transreid: Transformer-based object re-identification,''
\newblock in {\em Proceedings of the IEEE/CVF International Conference on Computer Vision}, 2021, pp. 15013--15022.

\bibitem{li2021diverse}
Yulin Li, Jianfeng He, Tianzhu Zhang, Xiang Liu, Yongdong Zhang, and Feng Wu,
\newblock ``Diverse part discovery: Occluded person re-identification with part-aware transformer,''
\newblock in {\em Proceedings of the IEEE/CVF Conference on Computer Vision and Pattern Recognition}, 2021, pp. 2898--2907.

\bibitem{zhu2022dual}
Haowei Zhu, Wenjing Ke, Dong Li, Ji~Liu, Lu~Tian, and Yi~Shan,
\newblock ``Dual cross-attention learning for fine-grained visual categorization and object re-identification,''
\newblock in {\em Proceedings of the IEEE/CVF Conference on Computer Vision and Pattern Recognition}, 2022, pp. 4692--4702.

\bibitem{zhu2023aaformer}
Kuan Zhu, Haiyun Guo, Shiliang Zhang, Yaowei Wang, Jing Liu, Jinqiao Wang, and Ming Tang,
\newblock ``Aaformer: Auto-aligned transformer for person re-identification,''
\newblock {\em IEEE Transactions on Neural Networks and Learning Systems}, 2023.

\bibitem{li2014deepreid}
Wei Li, Rui Zhao, Tong Xiao, and Xiaogang Wang,
\newblock ``Deepreid: Deep filter pairing neural network for person re-identification,''
\newblock in {\em Proceedings of the IEEE/CVF Conference on Computer Vision and Pattern Recognition}, 2014, pp. 152--159.

\bibitem{miao2019pose}
Jiaxu Miao, Yu~Wu, Ping Liu, Yuhang Ding, and Yi~Yang,
\newblock ``Pose-guided feature alignment for occluded person re-identification,''
\newblock in {\em Proceedings of the IEEE/CVF International Conference on Computer Vision}, 2019, pp. 542--551.

\bibitem{zhong2020random}
Zhun Zhong, Liang Zheng, Guoliang Kang, Shaozi Li, and Yi~Yang,
\newblock ``Random erasing data augmentation,''
\newblock in {\em Proceedings of the AAAI Conference on Artificial Intelligence}, 2020, vol.~34, pp. 13001--13008.

\bibitem{dao2023flashattention}
Tri Dao,
\newblock ``Flashattention-2: Faster attention with better parallelism and work partitioning,''
\newblock {\em arXiv preprint arXiv:2307.08691}, 2023.

\bibitem{liu2024survey}
Hanchao Liu, Wenyuan Xue, Yifei Chen, Dapeng Chen, Xiutian Zhao, Ke~Wang, Liping Hou, Rongjun Li, and Wei Peng,
\newblock ``A survey on hallucination in large vision-language models,''
\newblock {\em arXiv preprint arXiv:2402.00253}, 2024.

\end{thebibliography}
}

\end{document}